\documentclass[11pt]{article}
\usepackage[margin=1in]{geometry}
\usepackage[T1]{fontenc}
\usepackage{lmodern}
\usepackage{amsmath,amssymb,booktabs,graphicx,microtype}
\usepackage[hidelinks]{hyperref}
\hypersetup{pdftitle={Dynamic Shaping: Acquiring Interfaces and Constructing Execution over Frozen Pretrained Components},pdfauthor={Nathaniel Oh}}
\newcommand{\E}{\mathbb{E}}
\newcommand{\R}{\mathbb{R}}
\title{Dynamic Shaping:\\Acquiring Interfaces and Constructing Execution\\over Frozen Pretrained Components}
\author{Nathaniel Oh}
\date{November 15, 2026}
\begin{document}
\maketitle
\begin{abstract}
We introduce Dynamic Shaping, a method for constructing executable neural
arrangements through acquired response and partial-derivative interfaces over
frozen pretrained components. The method learns local descriptions of available
computation, ranks component rebindings using task-gradient alignment, transports
the current interface across a rebinding, and retains changes through behavioral
feedback. Continuous interface acquisition uses proximal policy optimization;
the source coefficients remain fixed. We derive an exact identity relating
transport error in component responses and partials to changes in action logits
and their sensitivities. A pixels-only ViZDoom Basic study evaluates first task
acquisition without a competent reference controller, demonstrations or intact
response targets. Across three initializations and 100 shared held-out game
seeds, initial arrangements succeed on 86/300 controller/episode pairs; joint
interface-and-binding construction succeeds on 295/300. Fixed-binding interface
acquisition also reaches 295/300. Each policy executes four source fragments,
with 110 acquired interface values and fixed random visual and action
projections. The artifact includes all candidate trials, interventions and costs;
all 1,200 recorded initial/retained outcomes replay exactly. The contribution
is an explicit profile--rebind--transport--retain procedure and a reproducible
reference-free acquisition case over frozen pretrained computation.
\end{abstract}

\noindent\textbf{Code and complete evidence:}
\url{https://github.com/calysteon/shaping-gap-benchmark}.\\
Checkpoint 8, immutable revision \texttt{73971c5a31f2acbfb7ea1e8131c456964e6e1826}.

\section{Introduction}
Pretraining produces numerical computation that can be used through interfaces
other than its original application. The research question here is how a
runtime can acquire an executable arrangement of that computation from a task
requirement and experience. We study a component-level construction process:
describe the effects of available frozen parts, choose how to connect them,
learn the interfaces through which they act, and retain a working execution.

We call this process \emph{Dynamic Shaping}. Its studied object is the acquired
arrangement and interfaces. A task contract supplies observable success
conditions, rather than a successful implementation. The runtime receives no
per-instance human repair or hand-selected successful binding. Frozen source
coefficients supply the available functions; task experience determines their
stimulus, binding and interface parameters.

The mechanism has four coupled operations. First, local response and partial
measurements describe the available components at states the runtime actually
visits. Second, task-gradient alignment ranks proposed changes to the active
bindings. Third, an affine transport changes the interface along with a binding
so that it can preserve useful responses and sensitivities. Fourth, real
behavioral feedback accepts or rejects the proposal. Standard policy
optimization continues acquiring the continuous interfaces between these
discrete construction steps.

Our contribution is the specific construction-and-transport procedure, its
exact transport-error identity, and an auditable first-acquisition experiment.
The procedure combines established tools: reinforcement learning, derivative
supervision, local approximation and affine alignment. Section~\ref{sec:related}
positions the combined method against the closest approaches. The experimental
question is whether it supplies reliable behavior absent from its initial
execution under the stated information and resource permissions.

We evaluate a simple pixels-only move-and-shoot task. Each initial runtime is
untrained on the task and occasionally succeeds, but falls below the specified
reliability contract. Acquisition receives game feedback without a competent
controller to imitate. This differs from the preceding reconstruction study,
which damaged a trained controller and supplied recorded intact responses
\cite{sg7}. The current study learns a first task implementation. Both the
full construction procedure and its fixed-binding interface control meet the
quality contract; the full procedure has no aggregate success advantage in
this experiment. The complete results and costs accompany the mechanism.

\section{Frozen substrate and acquired execution}
\label{sec:runtime}
Let $\Phi=\{\phi_1,\ldots,\phi_M\}$ be an inventory of callable scalar neural
components with immutable materialized parameters. A policy arrangement
$G=(G_1,\ldots,G_k)$ selects $k$ distinct identifiers. Continuous interfaces
determine the stimulus supplied to these components and how their responses
reach a fixed consumer. A shaping procedure receives $\Phi$, an initial
arrangement and interfaces, a behavioral contract $C$, permitted experience,
and a resource specification. It returns a retained executable arrangement,
interfaces and acquisition record.

The experimental inventory has $M=24$ RMSNorm/SwiGLU gate/up scalar responses
from the public Qwen3-0.6B-derived component bank, context 151003
\cite{sg5,artifact}. A fixed nine-dimensional source-coordinate compilation
materializes these responses. The original transformer attention graph,
scheduling, language head and native down projection are not executed by this
controller. Source coefficients and source normalization arrays stay fixed.
Appendix~\ref{app:native} specifies the executed scalar formula.

\begin{figure}[tb]
\centering\includegraphics[width=\linewidth]{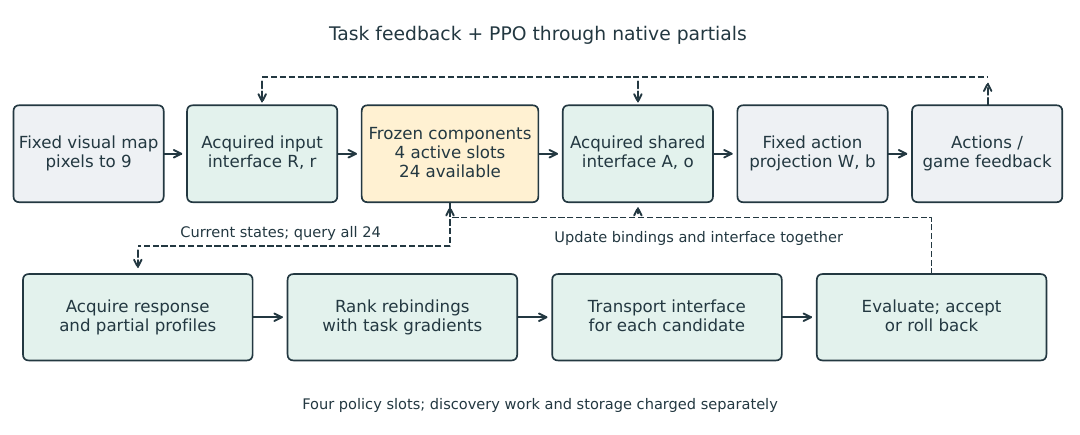}
\caption{Dynamic Shaping separates fixed computational parts from acquired
interfaces and bindings. Task experience drives continuous interface learning
and the discrete profile--rebind--transport--retain loop. Discovery queries
the inventory; policy execution uses four bound fragments.}
\label{fig:runtime}
\end{figure}

Each observation is a $160\times120$ RGB frame. Generic $16\times12$ block
means produce 576 numbers, standardized using unlabeled random-action
observations. A random fixed projection $E\in\R^{576\times9}$ and fixed
calibration produce a row vector $z\in\R^9$. Execution is
\begin{align}
 s &= \tanh(zR+r), & h &= (\phi_{G_1}(s),\ldots,\phi_{G_k}(s)),
 \label{eq:input}\\
 f &= hA+o, & \ell &= fW+b,\qquad \pi=\operatorname{softmax}(\ell).
 \label{eq:output}
\end{align}
Here $k=4$, $R\in\R^{9\times9}$, $r\in\R^9$, $A\in\R^{4\times4}$ and
$o\in\R^4$ are acquired: 110 real values. The random action projection
$W\in\R^{4\times3}$ and its zero bias $b$ remain fixed, as do $E$ and its
calibration. A four-value critic and scalar bias are auxiliary acquisition
state. Initially $R=0.15I$, $r=o=0$ and $A=I$. Generic diversity of native
partials at zero selects the initial bindings without task-performance
selection. Figure~\ref{fig:runtime} separates execution from acquisition.

The four-component limit is a \emph{policy execution} budget. Discovery queries
all 24 available components on its calibration states. Those queries,
temporary arrays, profile fits and candidate evaluations are charged
separately. Freezing the source does not freeze the complete policy: its
interfaces are learned parameters that can change end-to-end behavior.

\section{Response and partial interfaces for construction}
\label{sec:method}
\subsection{Continuous acquisition}
Proximal policy optimization (PPO) supplies the continuous optimizer
\cite{ppo}. Game rewards and generalized advantage estimates train the
input/output interfaces and auxiliary critic. No target action or target
hidden response is provided. Exact native partials connect the fixed functions
to the acquired stimulus. For action logit $a$, for example,
\begin{equation}
 \frac{\partial\ell_a}{\partial R_{ub}}
 =z_u(1-s_b^2)\sum_{j,v}
 \frac{\partial\phi_{G_j}}{\partial s_b} A_{jv}W_{va}.
 \label{eq:chain}
\end{equation}
Thus the learning signal can change how observations drive existing neural
responses while leaving their coefficients and the action consumer fixed.
Interface acquisition is training and is included in the ledger.

\subsection{Acquire an integrable local response description}
Every 20 updates, take 384 evenly spaced observations from the latest rollout
and recompute their current native input states $s_t$. Query responses and
analytic partials for all available components. Write normalized profile
coordinates as the column $q=((s-\mu)\oslash\sigma)^T$, with each coordinate
of $\sigma$ floored at 0.03.
For component $c$, fit
\begin{equation}
 \widehat\phi_c(q)=\alpha_c+L_cq+\tfrac12 q^TH_cq,
 \qquad \widehat J_c(q)=L_c+q^TH_c,
 \label{eq:profile}
\end{equation}
where $H_c$ is symmetric. Values and partials therefore describe one common
response function, rather than independent fitted fields. The first 256 states
fit the profile and the remaining 128 check approximation error.

The implementation uses ridge-regularized polynomial/Hermite regression with
both response and derivative observations, normalized per output. This is an
instance of established value-and-derivative supervision
\cite{sobolev,jacobian}. It yields a local description of a component's current
usable effects; the original component remains the executed function.

\subsection{Fit a value-and-partial transport}
For active slot $i$ and unused component $c$, let $y_t=\phi_{G_i}(s_t)$ be the
current slot response, $u_t=\widehat\phi_c(q_t)$ its proposed replacement,
$J^y_t=\partial_q\phi_{G_i}$ the measured current partial, and $J^u_t$ the
profile partial. The algorithm fits an affine bridge $a u+b$ by minimizing
\begin{equation}
 \E_t[(a u_t+b-y_t)^2]
 +\lambda\E_{t,d}[(aJ^u_{td}-J^y_{td})^2]+\kappa a^2,
 \label{eq:fit}
\end{equation}
with $\lambda=0.25$, $\kappa=10^{-6}$ and $a\in[-10,10]$. Here $\E_{t,d}$
averages over both states and native-input coordinates. Centering values gives
the implemented closed form
\begin{align}
 a&=\operatorname{clip}_{[-10,10]}
 \frac{\E_t[(u_t-\bar u)(y_t-\bar y)]+\lambda\E_{t,d}[J^u_{td}J^y_{td}]}
 {\E_t[(u_t-\bar u)^2]+\lambda\E_{t,d}[(J^u_{td})^2]+\kappa},
 \label{eq:bridge}\\
 b&=\bar y-a\bar u.
\end{align}
These targets describe the runtime's \emph{own current computation}. They are
not saved responses of a competent teacher or intact predecessor.

\subsection{Rank, transport and retain}
Let $\widehat\delta_t=a u_t+b-y_t$ and $v_i=A_{i,:}W$. The predicted raw-logit
displacement is $\Delta\widehat\ell_t=\widehat\delta_t v_i$. A policy-loss
gradient proxy $g_{ti}$, computed from rollout actions and standardized
advantages under the current probabilities, ranks each proposal by
\begin{equation}
 S_{i,c}=\sum_tg_{ti}\widehat\delta_t
 +0.05\,\frac12\E_t\!\left[
 \sum_a\pi_{ta}(\Delta\widehat\ell_{ta})^2
 -\left(\sum_a\pi_{ta}\Delta\widehat\ell_{ta}\right)^2\right].
 \label{eq:score}
\end{equation}
The second term is a softmax-curvature penalty. This score is a proposal
heuristic, not an exact PPO loss difference or a guarantee of reward gain.
Appendix~\ref{app:details} gives its implementation correspondence.

Candidates are sorted by $S_{i,c}$, followed by partial-transport error and
deterministic identifiers. For each of the top three, clone the current
interfaces and apply, using the old row $A_{i,:}$,
\begin{equation}
 G_i\leftarrow c,\qquad
 A_{i,:}\leftarrow a A_{i,:},\qquad
 o\leftarrow o+b A_{i,:}^{\rm old}.
 \label{eq:transport}
\end{equation}
Real development episodes evaluate these transported constructions. Accept
only a strict lexicographic improvement in success rate and mean return over
the unmodified current policy; otherwise roll back. Acceptance resets optimizer
moments. Continuous acquisition then resumes. A separate development score
selects the best snapshot across the complete acquisition horizon for final
retention. No terminal outcome selects any binding or interface.

\begin{center}
\fbox{\begin{minipage}{0.94\linewidth}
\small\textbf{Algorithm 1: Dynamic Shaping over a frozen inventory}
\begin{enumerate}
\item Generate the initial arrangement and fixed visual/action projections;
      calibrate observations without demonstrations.
\item Collect a rollout from the current policy; update input/output
      interfaces and critic using PPO through native partials.
\item At a construction round, measure available responses/partials and fit
      local profiles using Eq.~\eqref{eq:profile}.
\item For every slot/unused component pair, fit Eq.~\eqref{eq:bridge}, rank
      with Eq.~\eqref{eq:score}, and construct the top three transported
      candidates using Eq.~\eqref{eq:transport}.
\item Evaluate candidates on development episodes. Accept the best strict
      success/return improvement, or retain the current execution unchanged.
\item Record every trial and cost. Select and seal the best development
      snapshot; evaluate the retained construction on fresh episodes.
\end{enumerate}
\end{minipage}}
\end{center}

\section{Exact effect of interface transport}
\label{sec:identity}
The transport step has a useful algebraic property independent of the profile
approximation and proposal ranking. Let $\ell$ and $\ell'$ denote raw action
logits before and after Eq.~\eqref{eq:transport}, with all other bindings and
arrays fixed. Define the \emph{actual} response discrepancy at a native input
as $\delta=a\phi_c+b-\phi_{G_i}$. Use an input-by-action Jacobian convention.
Then
\begin{align}
 \ell'-\ell&=\delta v_i,\qquad v_i=A_{i,:}^{\rm old}W,
 \label{eq:identity}\\
 \partial_q\ell'-\partial_q\ell
 &= (a\partial_q\phi_c-\partial_q\phi_{G_i})^T v_i.
 \label{eq:partialidentity}
\end{align}
\textit{Proof.} All unchanged slot contributions cancel. The changed feature
contribution is $a\phi_c A_{i,:}^{\rm old}+b A_{i,:}^{\rm old}$ instead of
$\phi_{G_i}A_{i,:}^{\rm old}$. Multiplication by the unchanged $W$ gives
Eq.~\eqref{eq:identity}; differentiation with fixed $a,b,A,W$ gives
Eq.~\eqref{eq:partialidentity}. Consequently,
\begin{align}
 \|\ell'-\ell\|_2&=|\delta|\,\|v_i\|_2,\\
 \|\partial_q\ell'-\partial_q\ell\|_F
 &=\|a\partial_q\phi_c-\partial_q\phi_{G_i}\|_2\,\|v_i\|_2.
 \label{eq:error}
\end{align}
Exact response matching preserves the policy at that input; matching partials
also preserves its local logit sensitivity. Approximate matching exposes the
remaining discrepancy through Eq.~\eqref{eq:error}. The identity explains
what the transport controls. It does not certify the fitted profile away from
its measured states or guarantee improved game behavior; behavioral evaluation
supplies the retention decision.

\section{First acquisition from a task contract}
\label{sec:experiment}
\subsection{Task, information and protocol}
ViZDoom Basic supplies a single target at a randomized horizontal location and
three actions: move left, move right and shoot \cite{vizdoom}. Pixels are the
only policy observations. Game rewards train the interfaces; game variables
record outcomes but are not policy inputs. The runtime receives the action
set, generic preprocessing, optimizer, construction grammar and budgets.
It receives no target coordinates, object labels, scripted control rules,
demonstrations, teacher actions or intact-response targets.

The behavioral contract is at least 80 successes among 100 terminal episodes.
Initial arrangements can sometimes succeed; the missing behavior is reliable
control. The prospective strict criterion additionally required the initial
arrangement to succeed at most 30/100 times. We retain and report both
conditions separately.

The no-acquisition reference is the exact initial arrangement. Two acquisition
arms share its arrays, random seed and continuous optimization settings:
\emph{interface acquisition} keeps the initial bindings fixed, while
\emph{joint construction} adds the profile, rebinding and transport loop.
Both are autonomous learning procedures. Matched PPO update counts do not
imply matched total interactions: the joint arm uses extra development
evaluations for candidate selection, charged in the ledger.

Each run performs 160 updates of 1,024 sampled action steps, six PPO epochs
per update, minibatches of 128, learning rate 0.003, discount 0.99, GAE
parameter 0.95, clipping 0.2 and entropy coefficient 0.01. Calibration uses
1,024 random-action observations. Construction occurs every 20 updates:
eight rounds, three candidate evaluations per round, 24 trials per joint run.

Calibration game seeds start at 810000; acquisition seeds start at 820000.
Twenty development seeds, 830000--830019, serve all development comparisons.
All choices for a controller are sealed before its terminal evaluation by a
separate process on seeds 980000--980099. Initialization/action-sampling seeds
are 81, 82 and 83. The same 100 terminal game seeds are used for all controllers;
totals are controller/episode pairs, not 300 different maps. Seed 81 joint is
the engineering pilot. The implementation was locked after its development
run and before any terminal evaluation; seeds 82/83 use it unchanged.

\subsection{Held-out outcomes}
\begin{table}[ht]
\centering
\begin{tabular}{lrrrr}
\toprule
Runtime & 81 & 82 & 83 & Total\\
\midrule
Initial arrangement & 27/100 & 33/100 & 26/100 & 86/300\\
Interface acquisition & 100/100 & 95/100 & 100/100 & 295/300\\
Joint construction & 99/100 & 97/100 & 99/100 & 295/300\\
\bottomrule
\end{tabular}
\caption{Successful terminal episodes. Paired acquisition arms begin with
identical initial arrays; their duplicated initial outcomes are counted once.}
\label{tab:outcomes}
\end{table}

\begin{figure}[ht]
\centering\includegraphics[width=\linewidth]{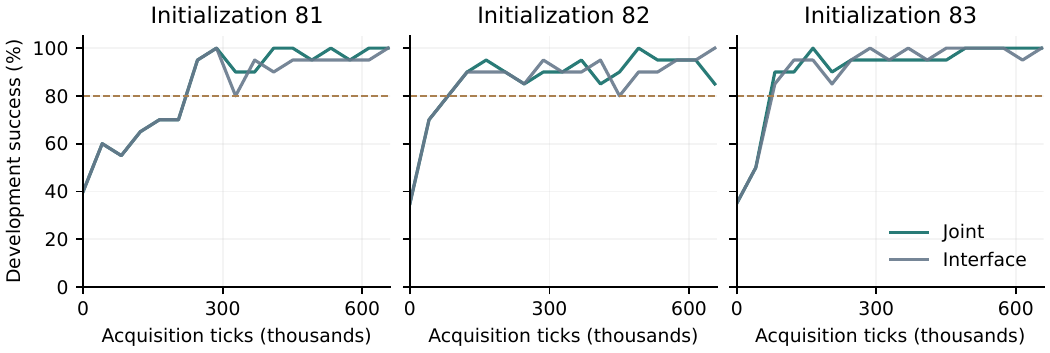}
\caption{Development success during acquisition. The dashed line is the
80\% contract. The horizontal axis counts continuous-acquisition game ticks;
calibration and development selection work are additional, not hidden in it.}
\label{fig:curves}
\end{figure}

Every acquired controller exceeds the 80/100 quality contract
(Table~\ref{tab:outcomes}). The strict initial-failure ceiling is met by
seeds 81 and 83 but missed by seed 82, whose initial score is 33/100. The
criterion is not revised after observing that result. Figure~\ref{fig:curves}
shows development trajectories; all terminal claims use the sealed snapshots.

Joint construction accepts 3, 2 and 5 rebindings during the three runs. Its
retained snapshots occur at updates 150, 120 and 160; their binding lists are
$(12,5,13,4)$, $(12,5,10,17)$ and $(21,23,3,2)$, versus the initial
$(12,5,11,17)$. Acceptance and final snapshot selection are distinct: a later
accepted change need not be in the best retained snapshot.

The two acquisition arms tie in aggregate success. Thus this experiment
establishes first acquisition under the stated permissions for both arms,
while leaving an aggregate advantage of discrete construction unresolved.
No random-weight substrate comparison or conventional game-agent performance
comparison is performed. The experiment does not isolate a causal benefit of
pretraining itself. Its positive result is a working executable construction
over the specified frozen pretrained inventory, without a successful task
reference.

\subsection{Post-acquisition interventions}
\begin{table}[ht]
\centering\small
\begin{tabular}{lrrrrrr}
\toprule
Intervention & J81 & I81 & J82 & I82 & J83 & I83\\
\midrule
Retained & 99 & 100 & 97 & 95 & 99 & 100\\
Original bindings & 71 & 100 & 97 & 95 & 24 & 100\\
Original input interface & 24 & 19 & 40 & 34 & 17 & 24\\
Original output interface & 36 & 97 & 82 & 66 & 7 & 95\\
Zero native outputs & 22 & 0 & 0 & 22 & 22 & 22\\
Zero standardized pixels & 0 & 0 & 22 & 22 & 22 & 0\\
\bottomrule
\end{tabular}
\caption{Successful episodes out of 100 after interventions without refitting.
J denotes joint construction and I denotes fixed-binding interface acquisition.}
\label{tab:interventions}
\end{table}

These diagnostics were declared after the seed-81 joint terminal result and
before running the interventions. They cause no further learning, selection or
retention changes. Original-bindings interventions are redundant for the
interface-only arm but remain recorded. Binding reversion has different effects:
it reduces J81 and J83 performance while leaving J82 success unchanged.
Input-interface reversion lowers every retained score. Source-output and
sensory interventions expose the remaining baseline behaviors rather than
assuming that every intervention must produce zero successes.

\subsection{Reproducibility and verification}
All 1,200 original initial/retained terminal episode records replay exactly in
fresh processes. This includes 300 duplicated initial records from paired
arms; the main experiment has 900 distinct initial/acquired-controller episode
pairs. The 3,000 intervention episodes are recorded separately. Fixed source,
projection and decoder arrays pass before/after hash checks. Centered finite
differences validate the learned-interface gradients to maximum absolute error
$2.55\times10^{-10}$.

A complete fresh acquisition from portable code and the compiled bank
reproduces every retained array, all eight binding-selection decisions and all
200 seed-81 initial/retained terminal episode records exactly. The public
artifact supplies the raw data, captured runners, profiles, ranked candidates,
rejections, selections, seals and manifests at a pinned Git revision
\cite{artifact}. Appendix~\ref{app:details} maps the paper to its entry points.

Language-model assistance from OpenAI Codex supported code development,
experiment orchestration, literature checks and manuscript preparation.
Runtime and package versions are included in the artifact.

\section{Acquisition and lifecycle costs}
\label{sec:costs}
\begin{table}[ht]
\centering\small
\begin{tabular}{lrrrr}
\toprule
Run & Development ticks & Seconds & RSS MiB & Accepted/tried\\
\midrule
81/joint & 78,424 & 58.21 & 64.52 & 3/24\\
81/interface & 29,060 & 53.32 & 62.12 & 0/0\\
82/joint & 79,688 & 59.29 & 66.59 & 2/24\\
82/interface & 27,708 & 51.54 & 62.28 & 0/0\\
83/joint & 44,844 & 56.98 & 67.20 & 5/24\\
83/interface & 20,440 & 53.33 & 63.78 & 0/0\\
\bottomrule
\end{tabular}
\caption{Complete original acquisition runs, including failed candidates and
acquisition after the best retained snapshot. Common additional work per row:
655,360 requested acquisition ticks and 4,096 calibration ticks.}
\label{tab:costs}
\end{table}

Each action repeats for four requested game ticks. Requested counts upper-bound
actual elapsed ticks when a terminal action ends early. Table~\ref{tab:costs}
includes discovery, continuous acquisition, profile construction, candidate
evaluation and development selection. Separate ledger entries charge terminal
tests, replay, interventions and fresh-acquisition verification. Source response
and partial-entry counters include the queried inventory, not only selected
parts. The records retain all rejected trials and all work performed after
the selected snapshot.

\begin{center}
\begin{minipage}{0.95\linewidth}
\centering
\textbf{Retained raw arrays}\par\smallskip
\begin{tabular}{lr}
\toprule
Arrays & Bytes\\
\midrule
Fixed projection, calibration and action decoder & 41,736\\
Pixel standardization & 9,216\\
Compiled source inventory & 5,336\\
Acquired input/output interfaces & 880\\
Auxiliary critic & 40\\
Active identifiers & 32\\
\midrule
Total & 57,240\\
\bottomrule
\end{tabular}
\end{minipage}
\end{center}
This is deployed array storage, not a full language model or whole-process RAM.
The 1,024 calibration observations temporarily occupy 4,718,592 raw bytes.
Profiles, candidate archives and logs are separate discovery/evidence storage;
they are not included in the deployment tally. Python peak RSS excludes the
separate game-engine process. Acquisition jobs ran sequentially, with brief
validation inference potentially overlapping; elapsed times describe this
machine, not a controlled throughput or electricity comparison. Source
pretraining and extraction are inherited costs, not assigned zero.

\section{Relation to existing methods}
\label{sec:related}
\paragraph{Reprogramming frozen computation.}
Adversarial reprogramming learns an input program that makes a fixed network
perform a different task \cite{reprogramming}. Frozen Pretrained Transformers
adapt language-pretrained computation to other modalities while retaining the
core attention/feedforward weights \cite{fpt}. Borrowed Geometry studies
cross-distribution head importance and cross-modal behavior through thin
interfaces over frozen text-pretrained slices \cite{borrowed}. These works
establish that task acquisition through interfaces over existing computation
is an important prior foundation. Here the acquired object additionally
includes discrete bindings of callable scalar components outside the original
transformer execution graph, with measured profiles guiding their replacement
and interface transport.

\paragraph{Pathways and stitching.}
PathNet evolves pathways while gradient learning updates permitted weights;
transfer freezes a learned path and evolves new paths \cite{pathnet}. Model
stitching learns a simple bridge between frozen neural representations
\cite{stitching}. Weight Agnostic Neural Networks demonstrate structural search
for task behavior while sharing a common weight value across connections
\cite{wann}. Dynamic Shaping combines a local construction proposal with
an explicitly transported interface, over an inventory whose source coefficients
all remain fixed. Current response/partial measurements determine a candidate
bridge; task-gradient alignment prioritizes candidates; feedback retains the
execution/interface pair. We study this complete loop with fixed visual/action
consumers and separately charged discovery work.

\paragraph{Derivative supervision and optimization.}
Sobolev training and Jacobian matching use derivative targets alongside values
\cite{sobolev,jacobian}. Our quadratic profiles and transport fit use these
established principles to describe available parts and align their local
effects. PPO supplies continuous policy optimization \cite{ppo}. The proposed
contribution is the specified combination of profile acquisition, task-directed
rebinding, interface transport and behavioral retention. It is not a claim of
inventing the constituent calculus, gradient optimizer or frozen-feature
learning paradigm.

\section{Discussion and future work}
The first-acquisition result makes a component runtime's learning cycle
externally observable: a task requirement and new experience lead to a retained
execution without a successful reference implementation. The source parameters
remain useful executable material throughout acquisition, while new interfaces
determine how that material is driven and read. The transport identity makes
one construction step mathematically explicit and separates its approximate
profile prediction from the actual executed effect.

The present evidence concerns one simple control scenario, one source family
and three initializations. It supplies no broad language-model improvement,
smaller-versus-larger model comparison or unbounded-depth result. The 110
interface values are learned parameters, and both acquisition arms use
substantial game experience. Fixed bindings already suffice for the same
aggregate success here, with less selection work. A matched-information study
of alternative proposal rules would isolate the incremental value of
response/partial-guided construction.

The next substantive question is continuing acquisition: whether retained
interfaces and construction knowledge reduce the cost of a second distinct
behavior while preserving the first. That requires charging discovery,
storage, execution, maintenance and failed reorganizations across the whole
sequence. Further work can test richer construction grammars, source families,
application contracts and deployment constraints. Sustainable deployment is a
motivation for improving the return on existing computation; measured resource
and energy savings require their own application studies.

\section{Conclusion}
Dynamic Shaping specifies how a runtime can acquire response/partial interfaces,
propose component rebindings, transport their interfaces and retain working
execution over fixed pretrained parts. Its transport step has an exact
output/sensitivity identity. The supplied game-control study demonstrates first
acquisition without a competent reference, with complete controls, exact replay
and lifecycle accounting. This establishes a concrete method and experimental
foundation for studying subsequent acquisition and reuse.

\clearpage
\appendix
\section{Executed scalar response and native partials}
\label{app:native}
For component $j$, the supplied compiled arrays define
\begin{align}
 \rho_j(s)&=\sqrt{\max(n_0+2s n_1+sQs^T+\epsilon_j,10^{-20})},\\
 g_j(s)&=(g_{0j}+sG_j^T)/\rho_j(s),\qquad
 u_j(s)=(u_{0j}+sU_j^T)/\rho_j(s),\\
 F_j(s)&=g_j(s)\,\operatorname{sigmoid}(g_j(s))\,u_j(s),\\
 \phi_j(s)&=\tanh\!\left((F_j(s)-m_j)/d_j\right).
\end{align}
$m_j,d_j$ are fixed source centering/scaling arrays; $Q$ is the compiled
symmetric normalization geometry. The source bank, including this
normalization, is immutable. On the positive, unclamped branch,
\begin{align}
 \partial_{s_b}\rho_j&=(n_{1b}+(sQ)_b)/\rho_j,\\
 \partial_{s_b}g_j&=(G_{jb}-g_j\partial_{s_b}\rho_j)/\rho_j,\\
 \partial_{s_b}u_j&=(U_{jb}-u_j\partial_{s_b}\rho_j)/\rho_j,\\
 \partial_{s_b}F_j
 &=[\sigma(g_j)+g_j\sigma(g_j)(1-\sigma(g_j))]u_j\partial_{s_b}g_j
   +g_j\sigma(g_j)\partial_{s_b}u_j,\\
 \partial_{s_b}\phi_j&=(1-\phi_j^2)\partial_{s_b}F_j/d_j.
\end{align}
The implementation clips the sigmoid exponential argument to $[-60,60]$ for
numerical stability. Its recorded gradient checks validate the exercised
numerical regime. Native source-coordinate fidelity is inherited from the
released materialization and reconstruction checks \cite{sg5,sg7}; the present
experiment adds interface-gradient checks, immutability checks and replay.

\clearpage
\section{Implementation correspondence and acquisition details}
\label{app:details}
The pinned Checkpoint 8 artifact has five canonical implementation files in
its \path{code/} directory: \path{construct.py}, \path{effects.py},
\path{substrate.py}, \path{environment.py} and \path{verify.py}.
In \path{construct.py}, the forward method of \texttt{Construction} implements
Eqs.~\eqref{eq:input}--\eqref{eq:output}. The gradient and update routines
implement continuous acquisition. The proposal routine implements
Eqs.~\eqref{eq:bridge}--\eqref{eq:score}; the rebinding routine implements
Eq.~\eqref{eq:transport}. The training entry point applies development
acceptance and seals selected arrays. Terminal testing runs after that seal;
verification performs no fitting.

For Eq.~\eqref{eq:score}, the rollout supplies actions $a_t$ and advantage
estimates. Advantages are standardized again on the 384 profiling states.
With current post-update probabilities, define the row vector
\begin{equation}
 d_t=-\frac{\widehat{\mathcal A}_t}{N}
       (\operatorname{onehot}(a_t)-\pi_t),\qquad
 g_{t,:}=(d_tW^T)A^T.
\end{equation}
This is the likelihood-gradient proxy used for proposals, rather than the
clipped PPO gradient or an unbiased new-policy rollout estimate. The recorded
rollout precedes the six continuous-optimization epochs. The score is used
only to allocate candidate evaluations; the actual game feedback decides
acceptance. Each round ranks $4(24-4)=80$ slot/component proposals and
evaluates three.

The quadratic profile has one constant, nine linear, nine diagonal-quadratic
and 36 cross terms per component. Response labels are centered and scaled
with standard deviation floored at $10^{-4}$; partial labels use the same
output scale. Response normal equations average over 256 states; derivative
normal equations average over $256\times9$ entries. Ridge $10^{-4}$ applies
to nonconstant coefficients. The same coefficients generate values and
partials. Check-point response RMSE across the 24 joint-run profile rounds
ranges approximately 0.060--0.181, and normalized-input partial RMSE
0.051--0.163. These are local approximation diagnostics, not task-success
certificates or checks on independently drawn game episodes.

Development acceptance maximizes the pair (success rate, mean return), using
candidate index to break ties, and requires improvement over the current
pair. Across checkpoints, final snapshot selection instead maximizes
success rate plus $10^{-4}$ times mean return. These two rules are deliberately
reported separately. All selections use the same development partition;
there is no statistical guarantee that a development improvement generalizes.

The artifact's \path{results.json} contains all six controller outcomes and
per-run costs. The separate \path{cost-ledger.json} adds terminal and diagnostic
work, and \path{MANIFEST.json} hashes the included files. The study distinguishes
policy array bytes, source inventory, temporary calibration arrays, discovery
archives and process RSS. Replaying or regenerating an acquisition is
verification work, not counted as an extra independent controller seed.
\end{document}